\documentclass[
]{ceurart}

\usepackage{listings}
\usepackage{xspace}
\usepackage{amsthm}
\usepackage{amssymb}
\usepackage{amsmath}
\usepackage{tikz}
\usetikzlibrary{positioning, shapes, arrows.meta}

\usepackage{xcolor}
\usepackage{draftwatermark}

\SetWatermarkText{To appear in: Proceedings of CILC-2025 (40th Italian Conference on Computational Logic, June 25–27, 2025, Alghero, Italy), \url{https://cilc2025.github.io/}}
\SetWatermarkScale{0.07}
\SetWatermarkColor[gray]{0.6}
\SetWatermarkAngle{270}
\SetWatermarkHorCenter{0.95\paperwidth}
\SetWatermarkVerCenter{0.5\paperheight}

\newcommand{\contrary}[1][\phantom{x}]{\overline{#1}}

\renewcommand{\L}{\mathcal{L}}

\newcommand{\R}{\mathcal{R}}
\newcommand{\ass}{\mathcal{A}}
\renewcommand{\arg}{\textit{Arg}}
\newcommand{\att}{\textit{Att}}

\theoremstyle{definition}
\newtheorem{definition}{Definition}[section]
\newtheorem{example}[definition]{Example}

\theoremstyle{plain}

\theoremstyle{remark}

\begin{document}

\copyrightyear{2025}
\copyrightclause{Copyright for this paper by its authors.
  Use permitted under Creative Commons License Attribution 4.0
  International (CC BY 4.0).}

\conference{40th Italian Conference on Computational Logic,
  June 25--27, 2025, Alghero, Italy}

\title{Weighted Assumption Based Argumentation to reason about ethical principles and actions}

\author[1]{Paolo Baldi}[%
orcid=0000-0003-2657-753X,
email=paolo.baldi@unisalento.it,
url=https://sites.google.com/view/paolobaldi,
]
\address[1]{Dept. of Human Studies, University of Salento, Lecce, Italy}

\author[1]{Fabio Aurelio D'Asaro}[%
orcid=0000-0002-2958-3874,
email=fabioaurelio.dasaro@unisalento.it,
url=https://sites.google.com/view/fdasaro,
]
\cormark[1]

\author[2]{Abeer Dyoub}[%
orcid=0000-0003-0329-2419,
email=abeer.dyoub@uniba.it,
url=https://www.abeerdyoub.com,
]
\address[2]{Dept. of Informatics, University of Bari ``Aldo Moro'', Via E. Orabona 4, Bari, 70125, Italy}

\author[2,3]{Francesca A. Lisi}[%
orcid=0000-0001-5414-5844,
email=francescaalessandra.Lisi@uniba.it,
url=https://www.uniba.it/it/docenti/lisi-francesca-alessandra,
]
\address[3]{Centro Interdipartimentale di Logica e Applicazioni (CILA), University of Bari ``Aldo Moro'', Via E. Orabona 4, Bari, 70125, Italy}

\cortext[1]{Corresponding author.}

\begin{abstract}
We augment Assumption Based Argumentation (\textit{ABA} for short) with weighted argumentation. In a nutshell, 
we assign weights to arguments 
and then derive the weight of attacks between ABA arguments. We illustrate our proposal through running examples in the field of ethical reasoning, and present an implementation based on Answer Set Programming.
\end{abstract}


\maketitle

\section{Introduction}
\label{intro}




Formal argumentation frameworks model reasoning with conflicting claims, based on the crucial notion of \emph{attacks} among arguments. These attacks are rendered as directed edges connecting nodes in a graph, while the arguments themselves are represented simply as nodes of the graph, see e.g. the seminal paper by Dung \cite{Dung1995}. 
Subsequent approaches, belonging to the field of \emph{structured argumentation}, see e.g. \cite{VanEemeren2018}, provide a more fine-grained representation of the argument nodes, equipping them with a logical structure, and using such structure for deriving from logical principles the occurrence of attacks among arguments. 

Assumption-based argumentation (see, e.g., \cite[Chapter 7]{VanEemeren2018}) is a prominent approach to structured argumentation. It represents arguments as logical derivations, built on the basis of two ingredients: \textit{rules}, which are considered to be non-defeasible, and \textit{assumptions}, which are taken instead to be the defeasible part of the argument, and possibly the target of attacks.


In this work, we introduce \emph{Weighted Assumption Based Argumentation frameworks (wABAs)}, which enrich ABA with \emph{weighted} arguments. These weights on arguments determine in turn  weights on the attacks among arguments, on the model of weighted abstract argumentation, see, e.g., \cite[Chapter 6]{Gabbay2021}.
This has several modeling advantages. 
On the one hand, the introduction of weights allows us to import ideas from fuzzy logic. On the other hand, since weighted arguments translate into weighted attacks, we may then allow for certain forms of incoherence in the semantics, making use of standard techniques in weighted abstract argumentation.

We propose an implementation of wABA using a translation into Answer Set Programming (ASP) by means of \texttt{clingo} answer set grounder and solver, and demonstrate the formalism and its implementation on a scenario involving AI ethics.

Our motivations for the introduction and implementation of wABA originates indeed from the general aim of addressing computational reasoning with ethical principles in AI, and more specifically in the domain of \textit{symbiotic human-AI interactions}. Deontological ethics, with its rule-based approach seems at first most well- suited for computational implementation in our setting.  
However, eliciting precise rules from general principles is far from obvious, and ethical rules can easily result in inconsistent outcomes.

We argue that, due to its peculiar features, wABA offers adequately expressive computational machinery both for the representation of ethical reasoning, and guidance on conflict resolution. 

Concerning representation, our approach  distinguishes, within wABA arguments, among (i) the general ethical principles, which play the role of the (defeasible) assumptions in the framework, (ii) the factual elements, which may be used as premises of arguments in addition to the assumptions, and (iii) the prescriptions of courses of actions, which appear as conclusions of arguments.

Arguments in wABA are thus well-suited to render \textit{prima-facie duties} \cite{ross2002right}, i.e., duties based on ethical principles, that may lead to conflicting prescriptions, and may be overriden on the basis of application context. 

Conflicts between ethically motivated prescriptions are naturally derived in wABA as attacks among the arguments. Following the \emph{extension}-based semantics of formal argumentation, wABA then allows one to represent various possible solutions to the ethical conflicts, i.e., different horns of ethical dilemmas, as different extensions. In the spirit of symbiotic AI \cite{Carnevale2024}, we take it to be a crucial feature of our approach that the AI agent does not substitute itself to the human agent, but disentangles, rather than solve, the ethical conflicts.

At the same time, wABA also offers guidance to the resolution of conflicts, due to the use of weights. Our design choice here is to avoid any a priori weighting of ethical principles, rather allowing the assignment of weights only to formulas standing for factual aspects, i.e. for the assessment of the context of application. In this respect, we extend and incorporate the fuzzy rule-based system developed in \cite{Dyoub2024} within our framework.
Only secondarily, on the basis of suitable computations, weights are carried over to arguments, then to attacks, and finally to the extensions of the argumentation framework. 
The user is thus ultimately confronted with weighted extensions, i.e., weighted solutions to ethical dilemmas, and can thus evaluate how strong a violation of certain ethical assumption she is willing to accept. 

The rest of the paper is structured as follows. In Section 2 we provide some background on abstract and assumption-based argumentation. In Section 3 we introduce wABA and in Section 4 we discuss our ASP implementation in \texttt{clingo}. Section 5 discusses the application of the framework to ethical reasoning and Section 6 the related work on computational approaches to ethical reasoning. Section 7 concludes the paper by wrapping up our contribution while hinting at future developments of the present work.




\section{Background}

In this section, we briefly recall the fundamental concepts of abstract argumentation frameworks (AAFs), assumption-based argumentation (ABA), and weighted abstract argumentation frameworks (wAAFs). These frameworks provide the  machinery upon which our proposed approach is constructed.

\subsection{Abstract Argumentation Frameworks}

We begin with Dung's abstract argumentation frameworks \cite{Dung1995}, which provide an abstract characterization of argumentation.

\begin{definition}[Abstract Argumentation Framework]
An \emph{abstract argumentation framework} (AAF) is a pair $(\arg,\att)$ where:
\begin{itemize}
    \item $\arg$ is a finite set of arguments;
    \item $\att \subseteq \arg \times \arg$ is a binary relation representing attacks between arguments.
\end{itemize}
\end{definition}

Given an AAF $(\arg, \att)$, for any $a, b \in \arg$, the notation $(a, b) \in \att$ indicates that $a$ attacks $b$. {We say that a set of  arguments $B\subseteq \arg$ \emph{defends} an  argument $a\in \arg $ if for each argument $b\in \arg$ that attacks $a$, there exists an argument $c\in {B}$ such that $c$ attacks $b$}. A subset ${B} \subseteq \arg$ is:
\begin{itemize}
    \item \emph{conflict-free} if there are no $a, b \in B$ such that $(a, b) \in \att$;
    \item \emph{admissible} if it is conflict-free and defends all its elements; 
    \item \emph{preferred} if it is a maximal (w.r.t. set inclusion) admissible set;
    \item \emph{grounded} if it is the least (w.r.t. set inclusion) admissible set;
    \item \emph{stable} if it is conflict-free and attacks every argument not in the set.
\end{itemize}

In addition to the semantics based on admissibility, alternative semantics have been introduced to capture different intuitions, particularly when admissibility leads to overly restrictive or unintuitive outcomes. 
These include the following:

\begin{itemize}
    \item \emph{Naive} extensions are the maximal (w.r.t. set inclusion) conflict-free subsets of $\arg$. Unlike admissible extensions, naive extensions do not require defense against attacks, focusing instead on maximizing conflict-freeness alone.
    
    \item \emph{Semi-stable} extensions are admissible sets whose range---i.e., the union of the set and the arguments it attacks---is maximal (w.r.t. set inclusion) among admissible sets. These extensions aim to approximate stable extensions when the latter do not exist.

    \item \emph{Stage} extensions are conflict-free sets whose range is maximal among all conflict-free sets. Like semi-stable semantics, they emphasize coverage (via attack) of the argument space, but do not require admissibility.

\end{itemize}

These non-admissibility-based semantics are particularly relevant in weighted or resource-bounded settings, where defense may be impractical or where broader coverage is desirable despite some lack of coherence.
In such cases, naive and stage extensions may yield more informative or robust outcomes than traditional admissible-based semantics.

\begin{figure}
\centering
\begin{tikzpicture}[
    node distance=1cm and 1cm,
    every node/.style={align=center, minimum width=3cm, minimum height=1cm},
    arrow/.style={->, thick}
]

\node (cf) {Conflict-Free};
\node[below left=of cf] (adm) {Admissible};
\node[below=of adm] (comp) {Complete};
\node[below=of comp] (pref) {Preferred};
\node[below=of pref] (ground) {Grounded};

\node[below right=of cf] (naive) {Naive};
\node[below=of naive] (stage) {Stage};
\node[below=of stage] (sst) {Semi-Stable};
\node[below=of sst] (stable) {Stable};

\draw[arrow] (ground) -- (pref);
\draw[arrow] (pref) -- (comp);
\draw[arrow] (comp) -- (adm);
\draw[arrow] (adm) -- (cf);

\draw[arrow] (stable) -- (sst);
\draw[arrow] (sst) -- (stage);
\draw[arrow] (stage) -- (naive);
\draw[arrow] (naive) -- (cf);

\draw[arrow] (stable) -- (pref);
\draw[arrow] (sst) -- (pref); 
\draw[arrow] (stage) -- (naive);

\end{tikzpicture}
\caption{Set-theoretic inclusion relations among semantics. All arrows denote subset inclusion.}
\end{figure}
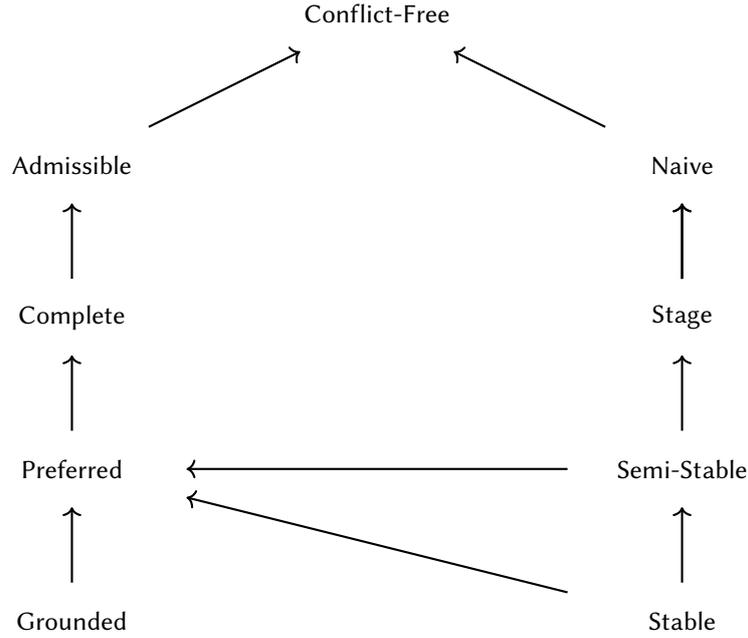

\subsection{Assumption-Based Argumentation}

Assumption-Based Argumentation (\textit{ABA}; see, e.g., \cite{Toni2014} for a primer) provides a structured argumentation framework grounded in a deductive system, based on rules and defeasible assumptions.

\begin{definition}[ABA Framework]
An \emph{assumption-based argumentation} (ABA) framework is a tuple $( \mathcal{L}, \R, \ass, \contrary)$ where:
\begin{itemize}
    \item $\mathcal{L}$ is a formal language;
    \item $\R$ is a set of inference rules of the form $\phi \leftarrow \phi_1, \dots, \phi_n$ with $\phi, \phi_i \in \mathcal{L}$;
    \item $\ass \subseteq \mathcal{L}$ is a non-empty set of assumptions;
    \item $\contrary: \ass \to \mathcal{L}$ is a total mapping that assigns to each assumption its contrary.
\end{itemize}
An ABA is said to be \textit{flat} if and only if assumptions only appear in the body of rules.
\end{definition}

Henceforth, for any rule $r\in R$ of the form $\phi \leftarrow \phi_1, \dots, \phi_n \in \R$ we let $body(r) = \{  \phi_1, \dots, \phi_n\}$ and $head(r) = \phi$. Arguments in ABA are deductions, denoted by $\Phi\vdash\psi$, where $\Phi$ is a set of assumptions and $\psi$ is any formula in $\mathcal{L}$, obtained from $\Phi$ by applying one or more rules. Given an ABA argument $x$ (i.e. a deduction), we denote by ${rules(x)}$ the set of rules \textit{supporting} it.
Attacks are defined via contraries. 
Specifically, we will have that an argument $x$ attacks an argument $y$, if and only if $x$ is a deduction of the form $\Phi'\vdash \overline{\psi}$, and 
$y$ is a deduction of the form $\Phi,\psi\vdash \phi$,  
where $\psi\in A, \phi\in \mathcal{L}$, $\overline{\psi}$ is the contrary of $\psi$ and $\emptyset\subseteq \Phi,\Phi'\subseteq A$. Any ABA framework thus naturally defines a corresponding abstract argumentation framework associated with it, from which extensions can be extracted.



\begin{figure}
    \centering
    \includegraphics[width=0.7\linewidth]{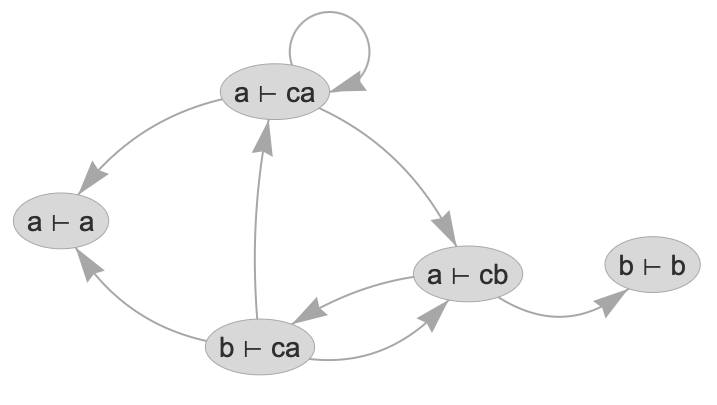}
    \caption{ABA framework visualization for Example \ref{example:ABA1}.}
    \label{fig:ABA1}
\end{figure}

\begin{example}\label{example:ABA1}
    As a refresher, consider the flat ABA framework consisting of \textit{atoms} $a$, $b$, $ca$, and $cb$, where $a$ and $b$ are \textit{assumptions}, \textit{contraries} are given by $\overline{a}=\textit{ca}$ and $\overline{b}=\textit{cb}$, and \textit{rules} are $\textit{ca} \leftarrow a$, $\textit{ca} \leftarrow b$, $\textit{cb} \leftarrow a$. Recall that in ABA an argument is a deduction. Examples of arguments in this ABA are $b\vdash b$, $a \vdash \textit{ca}$, and $b\vdash \textit{ca}$. Attacks are then derived by considering contraries: for example, $b \vdash \textit{ca}$ attacks $a\vdash \textit{ca}$ as the derived atom $\textit{ca}$ is the contrary of the assumption $a$. The full framework is shown in Figure \ref{fig:ABA1}, which was produced with the \textit{PyArg}  library \cite{Borg2022}. This graph can be treated as a standard AAF. For instance, the only stable extension of this framework is $\{b\vdash ca, b\vdash b\}$.
\end{example}

\subsection{Weighted Abstract Argumentation}

Weighted extensions of abstract argumentation frameworks \cite{Dunne2011} assign numerical weights to arguments or attacks, enabling reasoning with preferences, strengths, or costs.

\begin{definition}[Weighted AAF]\label{def:wAAF}
A \emph{weighted abstract argumentation framework} (wAAF) is a triple $(\arg, \att, w)$ where:
\begin{itemize}
    \item $(\arg, \att)$ is an abstract argumentation framework;
    \item $w: \att \to \mathbb{R}^{+}$ is a function assigning a positive real-valued weight to each attack.
\end{itemize}
\end{definition}


An \textit{inconsistency budget} $\beta$ is typically used to specify a degree of inconsistency one is willing to tolerate in a given scenario. In other words, attacks that weigh up to $\beta$ may be discarded from the framework. Semantics are then defined with respect to the inconsistency budget $\beta$ and a standard semantics $\sigma$ of AAF, see, e.g., \cite{Bistarelli2021}. One usually refers to these semantics as $\beta$-$\sigma$ extensions, e.g., $3$-stable, $2$-admissible, $5$-grounded, etc. Note that whenever $\beta=0$ we obtain the standard argumentation semantics $\sigma$, e.g., $0$-stable is the standard AAF stable semantics, $0$-grounded is the standard AAF grounded semantics, etc. 

\begin{example}
    Consider the weighted abstract argumentation framework $(\arg, \att, w)$ where:
    \[
    \arg = \{a, b, c\}, \quad \att = \{(a, b),\ (b, c),\ (c, a)\}, \quad w((a, b)) = 2,\ w((b, c)) = 1,\ w((c, a)) = 4.
    \]
    This framework forms a directed cycle where each argument attacks one other. Under standard stable semantics (i.e., with $\beta = 0$ and $\sigma$=stable), there exists no extension, as each argument is attacked by another, and no conflict-free set can defend against all incoming attacks.

    Now consider the framework under a budgeted stable semantics with inconsistency budget $\beta = 3$. In this case, we may discard attacks whose cumulative weight does not exceed $3$. For instance, we may choose to discard the attack $(c, a)$, which has weight $4$, but this alone would exceed the budget. However, if we discard $(a, b)$ and $(b, c)$, whose combined weight is $3$, we remain within the budget.

    After discarding $(a, b)$ and $(b, c)$, the only remaining attack is $(c, a)$. The set $\{a, b\}$ is now conflict-free with respect to the reduced attack relation and attacks $c$, making it a valid $3$-stable extension.

    This example illustrates how the introduction of a budget $\beta$ permits certain extensions that are disallowed under classical semantics, thereby enabling reasoning in the presence of bounded inconsistency or uncertainty.
\end{example}

\section{Weighted Assumption Based Argumentation}

 In this Section we develop our proposed framework for Weighted Assumption Based Argumentation (\textit{wABA}). The basic idea is to extend ABA by assigning a cost to certain atoms, and use these costs to compute the weight of attacks.

Just like in plain ABA, arguments are defined as deductions.
Attacks are still defined in terms of contraries as in ABA, i.e. any argument of the form $\Phi,\psi\vdash \phi$ may only be attacked by arguments of the form $\Phi'\vdash \overline{\psi}$.

\begin{definition}[Weighted Assumption-Based Argumentation]\label{def:waba}
A \emph{wABA framework} is a tuple
\[(\mathcal{L},\R,\ass,\contrary,\mathcal{S},w)\]
where  $(\mathcal{L},\R,\ass,\contrary)$ is an ABA framework, $S\subseteq\mathbb{R}_0^+ \cup \{\infty\}$ (where $\mathbb{R}^+_0$ is the set of nonnegative real numbers) and $\mathcal{S} = (S,\oplus,\otimes,e_\oplus,e_\otimes)$ is a \textit{semiring}\footnote{This means that  $(S,\oplus,e_{\oplus})$ is a commutative monoid, $(S,\otimes,e_{\otimes})$ is a monoid, distributivity holds w.r.t. both operators, and any element is annihilated by $e_\oplus$. }.
The function $w\colon \mathcal{L} \to S $ is a \textit{weight} such that 
$w(\phi)= e_\otimes$ for each $\phi\in \ass$.

The weights are then extended to attacks $(x,y)$ among arguments $x$ and $y$ by letting\footnote{In other words, the weight of the attack is the weight of the attacking derivation, which is in turn computed by suitably aggregating the weights of the atoms occurring in the body of the rules supporting the derivation.}:
\[ w((x,y )) = \bigotimes_{r  \ \in\ \mathrm{rules}(x)}\  \bigotimes_{\phi \ \in \ \mathrm{body}(r)}  w(\phi).  \]
\end{definition}
As in weighted abstract argumentation, one can then choose to discard attacks that do not exceed an inconsistency budget $\beta \in S$, using the operation $\oplus$ of the semiring, as follows. 
\begin{definition}[$\beta$-$\sigma$ extensions]
Let 
$(\mathcal{L},\R,\ass,\contrary,\mathcal{S},w)$ be a WABA framework,  over the semiring $\mathcal{S} = (S,\oplus,\otimes,e_\oplus,e_\otimes)$, $\sigma$  be a set of extensions over the abstract argumentation framework 
$(\arg,\att)$ associated with 
$(\mathcal{L},\R,\ass,\contrary)$,  and $\beta \in S$.
   We say that a subset  $\arg'\subseteq \arg$  is a $\beta$-$\sigma$ extension for $(\mathcal{L},\R,\ass,\contrary,\mathcal{S},w)$ iff there is a subset $\att'\subseteq \att$ such that \[ \bigoplus_{(x,y) \in \att'} w(x,y) \leq \beta\] and $\arg'$ is a $\sigma$ extension for $(\arg,\att\setminus \att')$ .
\end{definition}
We say that a wABA is \emph{flat} if $(\L,\R,\ass,\contrary)$ is a flat ABA. In the following, we assume that all wABAs are flat, unless stated otherwise.

We illustrate the notions we have introduced with the following example.

\begin{figure}
    \centering
    \includegraphics[width=0.7\linewidth]{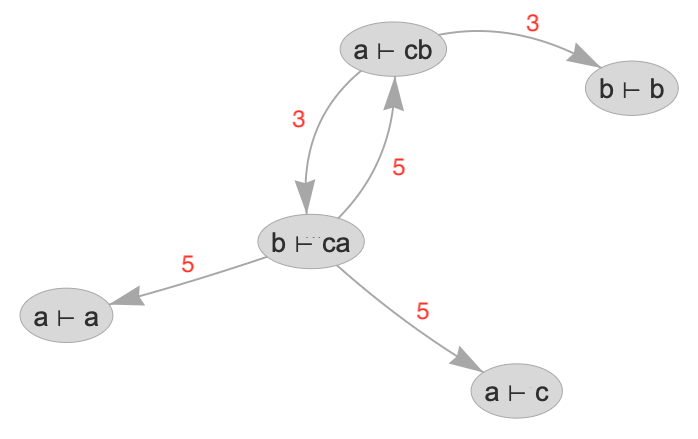}
    \caption{wABA framework from Example \ref{example:wABA1}. Weights of attacks are displayed in red. Sets of rules used in derivations are: $\textit{rules}(b\vdash ca) = \{ d\leftarrow\top,\ ca\leftarrow b,d \}$, $\textit{rules}(a\vdash cb)=\{ d\leftarrow\top,\ c\leftarrow a,d,\ cb\leftarrow a,c \}$, $\textit{rules}(a\vdash c)=\{ d\leftarrow\top,\ c\leftarrow a,d \}$, and $\textit{rules}(a\vdash a) = \textit{rules}(b\vdash b) =\emptyset$.}
    \label{fig:wABA1}
\end{figure}

\begin{example}\label{example:wABA1} Consider the wABA consisting of atoms $\{a, b, c, d, \textit{ca}, \textit{cb}\}$, where $a$ and $b$ are assumptions, $c$ and $d$ are contextual information, and contraries are defined via $\overline{a}=\textit{ca}$ and $\overline{b}=\textit{cb}$. Weight of contextual atoms are defined as $w(c)=3$ and $w(d)=5$. Rules are:
\begin{align*}
\textit{ca}\leftarrow b, d\\
\textit{cb} \leftarrow a, c\\
c \leftarrow a, d\\
d \leftarrow \top
\end{align*}
The resulting framework may be depicted as in Figure \ref{fig:wABA1}. Note that if one has an appropriate inconsistency budget then some pairs of attacks may be removed from the framework, e.g., the mutual attacks from $b\vdash \textit{ca}$ and $a\vdash cb$ for $\textit{rules}(b\vdash ca) = \{ d\leftarrow\top,\ ca\leftarrow b,d \}$ and $\textit{rules}(a\vdash cb)=\{ d\leftarrow\top,\ c\leftarrow a,d,\ cb\leftarrow a,c \}$. The resulting extensions can be calculated according to the standard AAF definition.
\end{example}

\section{Implementation}

Our implementation of (Weighted) Assumption-Based Argumentation is based on Answer Set Programming (ASP) using \texttt{clingo} \cite{Gebser2019}. The code is freely available at \url{https://github.com/dasaro/ABA-variants/tree/main/WABA}.

\subsection{Theoretical assumptions}
The current encoding supports \textit{context-free}, \textit{naive} and \textit{stable} semantics.

Our implementation adopts the \textit{min–max semiring}
\[
\mathcal{S} = (\mathbb{N}\cup\{\infty\},\max,\min,0,\infty)
\]  
as our default structure for wABA.  This choice is driven primarily by the inherently fuzzy nature of inputs, and \texttt{clingo} implmenetation, which provides native support for natural numbers but not for real values.
We start by detailing the translation procedure which translates a wABA framework into an answer set program, which is largely based on ASPforABA \cite{lehtonen2023aspforaba}.

\subsection{Translation}
Let $(\mathcal{L}, \R, \ass, \contrary, \mathcal{S}, w)$ be a flat wABA. For compactness and readability, in the following we use the standard \texttt{clingo} abbreviation ``\texttt{;}'' which unpacks an expression, e.g., of the form \texttt{r(a1,b1; $\dots$; an,bn)} into a set of clauses \texttt{r(a1,b1)}, $\dots$, \texttt{r(an,bn)}.

The set of assumptions $\ass=\{a_1$, $\dots$, $a_n\}$ gets translated to:
\begin{lstlisting}
assumption(a1; ...; an).
\end{lstlisting}

Each assumption is assigned to its contrary via the $\contrary$ operator, e.g., $\overline{a_1}=c_1$, $\dots$, $\overline{a_n}=c_n$. This gets translated to:
\begin{lstlisting}
contrary(a1,c1; ...; an,cn).
\end{lstlisting}

Each rule of the form $h\leftarrow b_1, \dots, b_m$, where $h \in \mathcal{L\setminus\ass}$ is the \textit{head}, and $b_1$, $\dots$, $b_m \in \mathcal{L}$ are the \textit{body} of the rule, is translated to:
\begin{lstlisting}
head(id,h). body(id,b1; ...; id,bm).
\end{lstlisting}
where \texttt{id} is a unique identifier assigned to the rule, so that different rules get assigned different identifiers.

% ...

Finally, weights for atoms in $\mathcal{L}$, e.g., $w(d_1)=w_1$, $\dots$, $w(d_l)=w_l$ translate to:
\begin{lstlisting}
weight(d1,w1; ...; dl, wl).
\end{lstlisting}

\subsection{Semantics}

The semantics file \texttt{core.lp} is fixed for all semantics and is described in what follows. We start by declaring a global inconsistency budget $\beta$:
\begin{lstlisting}
budget(beta).
\end{lstlisting}
This can be set from the Command Line by using clingo's \texttt{--const beta=N} in-built flag. Note that if one wants to enumerate all the extensions, regardless of their budget, \texttt{--const beta=\#sup} may be used.

Each assumption can be either {\tt in} or {\tt out} of a candidate extension:
\begin{lstlisting}
in(X)  :- assumption(X), not out(X).
out(X) :- assumption(X), not in(X).
\end{lstlisting}

Support propagates from selected assumptions through the rules in a bottom‐up fashion:
\begin{lstlisting}
supported(X) :- assumption(X), in(X).
supported(X) :- head(R,X), triggered_by_in(R).
triggered_by_in(R) :- head(R,_), supported(X) : body(R,X).
\end{lstlisting}
We then assign supported atoms a weight according to the rules they were produced from, using the $\otimes = \min$ operator (and note that assumptions are assigned weight $e_{\otimes}=\infty$ which in clingo is rendered as the constant \texttt{\#sup}):
\begin{lstlisting}
supported_with_weight(X,#sup) :- assumption(X), in(X).
supported_with_weight(X,W) :- supported(X), weight(X,W).
supported_with_weight(X,W) :-
    supported(X), head(R,X),
    W = #min{ V, B : body(R,B), supported_with_weight(B,V) }.
\end{lstlisting}

Attacks arise from contraries:
\begin{lstlisting}
attacks_with_weight(X,Y,W) :-
    supported(X), supported_with_weight(X,W),
    assumption(Y), contrary(Y,X).
\end{lstlisting}

As usual in wAAFs, we may choose to discard any subset of those attacks, paying their full weight; the total discarded weight must not exceed the budget w.r.t. semiring operation $\oplus=\max$:
\begin{lstlisting}
{ discarded_attack(X,Y,W) : attacks_with_weight(X,Y,W) }.
extension_cost(C) :- C = #max{ W, X, Y : discarded_attack(X,Y,W) }.
:- extension_cost(C), C > B, budget(B).
\end{lstlisting}

Since we do not want discarded attacks to be effective, we also introduce the notion of a successful attack:

\begin{lstlisting}
attacks_successfully_with_weight(X,Y,W) :-
    attacks_with_weight(X,Y,W), not discarded_attack(X,Y,W).
\end{lstlisting}

Once we have figured out how arguments are built from the assumptions, including their weights, contraries and discarded attacks according to the budget, we are left with the task of selecting an appropriate semantics to choose valid extensions from.

First, we introduce useful shorthands:
\begin{lstlisting}
defeated(X) :- attacks_successfully_with_weight(_,X,_).
not_defended(X) :- attacks_successfully_with_weight(Y,X,_), not defeated(Y).
\end{lstlisting}
which state that an atom is \textit{defeated} iff it receives a successful attack, and that it is \textit{not defended} iff it has an undefeated attacker. These shorthands will help simplify the form of semantics below.

Then, we define four popular semantics:
\paragraph{Conflict-Free}
Conflict freeness ensures that, once discarded attacks are removed in the way outlined above, two assumptions in the same extension do not attack each other:
\begin{lstlisting}
:- in(X), defeated(X).
\end{lstlisting}

\paragraph{Admissible semantics} Admissibility is a widely adopted property of argumentation frameworks, and many other semantics (such as the stable semantics defined below) produce subsets of admissible sets. It is implemented by making sure it is context-free and all its elements are defended:
\begin{lstlisting}
:- in(X), defeated(X). % conflict-freeness
:- in(X), not_defended(X).
\end{lstlisting}

\paragraph{Stable Semantics} 
In the stable semantics, every (conflict-free) assumption outside the extension must be defeated by a non‐discarded attack:
\begin{lstlisting}
:- in(X), defeated(X). % conflict-freeness
:- out(X), not defeated(X).
\end{lstlisting}

\paragraph{Naive Semantics} The naive semantics is a maximal conflict-free set w.r.t. to set inclusion, which is not necessarily admissible:
\begin{lstlisting}
:- in(X), defeated(X). % conflict freeness
#heuristic in(X) : assumption(X). [1,true]
\end{lstlisting}
which ensures the maximal set of assumptions is ``in'' regardless of the extension being admissible or not.

\section{A Weighted ABA Approach to Ethical Reasoning}

In this section, we apply the proposed \texttt{clingo} implementation of wABA to ethical reasoning  involving conflicting principles and contextual medical information. We begin by revisiting the approach proposed in \cite{Dyoub2024}, which we expand into a more comprehensive (W)ABA-based framework for ethical decision-making. \cite{Dyoub2024} focuses on the following motivating scenario:

\begin{example}[Patient Dilemma]\label{example:patientDilemma}
A care robot approaches a patient to administer medication, which would very likely address some health issues. The patient, however, refuses to take it. Should the robot attempt to persuade the patient, or should it respect the patient's decision?
\end{example}

The \textit{dilemma} involves a potential conflict between \emph{autonomy}, which prioritizes the patient's decision, and \emph{beneficence}, which emphasizes promoting the patient's well-being. Depending on the clinical and institutional context, other principles such as \emph{non-maleficence} (e.g., medication may have harmful side effects) and \emph{justice} (e.g., resource constraints across patients) may also be relevant.

\begin{figure}
	\centering
	\includegraphics[width=0.9\linewidth]{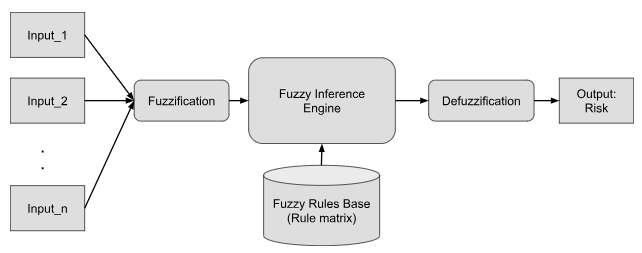}
	\caption{Architecture of the fuzzy system for ERA.}
	\label{fig:fuzzymodel}
\end{figure}

Figure \ref{fig:fuzzymodel} shows the architecture of the fuzzy logic based system for ethical risk assessment (ERA) proposed in \cite{Dyoub2024}. Dyoub \& Lisi used ERA system to assess the possible ethical risk of causing physical harm to the patient in the above mentioned Patient Dilemma. For evaluating the physical harm risk in this case, we can consider different parameters as inputs to ERA system such as, severity of health condition of the patient, mental/psychological condition of the patient, physiological indicators of well-being, etc. These inputs are rated on some scale (e.g. between 0 and 10). The crisp values are then fuzzified into fuzzy sets with linguistic variables. For example, the fuzzy set label for \textit{severity} could be one of $\{\textit{very\_low}, \textit{low}, \textit{medium}, \textit{high}, \textit{very\_high}\}$. Then, the fuzzy inference engine uses the fuzzy rules from the fuzzy rule base to calculate the risk level. For instance, a patient in a severe health condition ($90\%$) with reduced mental capacity ($80\%$) is evaluated as $\textit{very\_high}$ risk ($95\%$). These values may be readily piped in into our wABA framework.

\subsection{Plain ABA approach}
We first show how we can embed linguistic labels from the fuzzy system to reason about which ethical principles are satisfied (resp. violated).

\begin{example}[Patient Dilemma in ABA]\label{example:PatientDilemmaInABA} Let us assume that, on a scale from $0$ to $10$, a patient in a highly severe physical condition ($9/10$), high mental risk ($8/10$) and refusing to taking her medications, is evaluated as a $\textit{high\_risk}$ individual by the underlying fuzzy systems--meaning that not taking the medications could lead to developing severe physical consequences (including death).

We may formalize principles involved in this scenario through assumptions $\textit{respect\_autonomy}$ and $\textit{act\_beneficently}$. Other atoms in the language are $\textit{risk}(\textit{high})$, reflecting the fuzzy evaluation of risk, $\textit{action}(\textit{give\_meds})$ as the action of giving medicines, $\textit{action}(\textit{dont\_give\_meds})$ as the action of not giving medicines to the patient, $\textit{opinion}(\textit{refuses\_meds})$ as the contextual information that the patient refuses the medications, $\textit{ca}$ as the contrary of respect for autonomy (i.e., $\overline{\textit{respect\_autonomy}}=\textit{ca}$) and $\textit{cb}$ as the contrary of $\textit{act\_beneficently}$ (i.e., $\overline{\textit{act\_beneficently}}=\textit{cb}$). Experts define rules as follows:
\begin{align*}
    \textit{opinion}(\textit{refuses\_meds}) \leftarrow \top\\
    \textit{risk}(\textit{high})\leftarrow\top\\
    \textit{action(give\_meds)}\leftarrow \textit{risk}(\textit{high}),\ \textit{act\_beneficently}\\
    \textit{action(dont\_give\_meds)}\leftarrow \textit{opinion}(\textit{refuses\_meds}), \textit{respect\_autonomy}\\
    ca \leftarrow \textit{action}(\textit{give\_meds}),\ \textit{opinion}(\textit{refuses\_meds})\\
    cb \leftarrow \textit{action}(\textit{dont\_give\_meds}),\ \textit{risk}(\textit{high}) 
\end{align*}
whose interpretation is intuitive. We can calculate the stable extensions in our framework, namely $\{\textit{act\_beneficently}\}$, from which action \textit{give\_meds} follows, and
$\{\textit{respect\_autonomy}\}$, which supports \textit{not giving meds}. Note that, in an alternative situation where the patient does \textit{not} refuse to take the meds both ethical principles may be satisfied: indeed, in this scenario we get the unique extension $\{\textit{act\_beneficently,\ respect\_autonomy}\}$
from which one can derive that giving  medications satisfies both \textit{beneficence} and \textit{autonomy}, since the patient does not refuse to take the medications in the first place.
\end{example}

This simple example can be further extended by considering more nuanced interactions between ethical principles and their contraries. A fuller example, involving the other two standard principle of bioethics (namely, \textit{non maleficence} and \textit{justice}) is available on our GitHub repository, and the interested reader can try it out.

We now turn to discussing how such reasoning about ethical principles may be enriched with weights that allow for inconsistencies.

\subsection{wABA approach}

In addition to assumptions and contraries, wABA introduces \textit{weighted information}, and an \textit{inconsistency budget} $\beta$. In this way, we can model more nuanced aspects of ethical decision-making.

\begin{example}[Patient Dilemma in wABA] In order to show the characteristics of wABA, we further elaborate on Example \ref{example:PatientDilemmaInABA}. Recall that the patient is in a severe physical condition ($0.9$) with reduced mental capacity ($0.8$). In this case, we let the fuzzy component of our system \textit{defuzzify} its output, which provides us with a numeric value for the risk of phyisical harm. Let us assume the output defuzzified value for the risk is $0.7$. Furthermore, let us assume that the patient is very reluctanct to taking medications, which we assign (or the fuzzy system assigns) a weight of $0.9$. This amounts to saying that the contextual information  is described by:
\[ w(\textit{risk})=0.7,\quad w(\textit{opinion(refuses\_meds)})=0.9 \]
It is worth noting here that we have dropped the lingustic variables appearing in the first example (i.e., $\textit{risk}(\textit{high})$) in favor of explicit weights. We modify the rules to reflect this:

\begin{align*}
    \textit{opinion}(\textit{refuses\_meds}) \leftarrow \top\\
    \textit{risk}\leftarrow\top\\
    \textit{action(give\_meds)}\leftarrow \textit{risk},\ \textit{act\_beneficently}\\
    \textit{action(dont\_give\_meds)}\leftarrow \textit{opinion}(\textit{refuses\_meds}), \textit{respect\_autonomy}\\
    ca \leftarrow \textit{action}(\textit{give\_meds}),\ \textit{opinion}(\textit{refuses\_meds})\\
    cb \leftarrow \textit{action}(\textit{dont\_give\_meds}),\ \textit{risk}
\end{align*}
where the weights of $\textit{risk}$ and $\textit{refuses\_meds}$ automatically enter the computation by means of wABA semantics.

In the implementation, we use the appropriate predicate \texttt{weight/2} as follows:
\begin{lstlisting}
head(ctx1, risk).          weight(risk, 7).
head(ctx2, refuses_meds).  weight(refuses_meds, 9).
\end{lstlisting}

Assumptions, rules and contraries are implemented in the exact same way as in plain ABA (see Example \ref{example:PatientDilemmaInABA}). Note that if we let $\beta=0$ we reconstruct exactly the same extensions as in Example \ref{example:PatientDilemmaInABA}, i.e., the (postprocessed for readability) output looks as follows:
\begin{lstlisting}
Answer: 1
in(respect_autonomy) supported_with_weight(action(dont_give_meds),9) supported_with_weight(cb,7) extension_cost(0)

Answer: 2
in(act_beneficently) supported_with_weight(action(give_meds),7) supported_with_weight(ca,7) extension_cost(0)
\end{lstlisting}
However, if we list \textit{all} extensions, we get another \textit{inconsistent} stable extension resulting from the removal of the ethical inconsistency between \texttt{respect\_autonomy} and \texttt{act\_beneficently}:
\begin{lstlisting}
Answer: 3
in(respect_autonomy) in(act_beneficently) supported_with_weight(action(dont_give_meds),9) supported_with_weight(action(give_meds),7) supported_with_weight(ca,7) supported_with_weight(cb,7) extension_cost(7)
\end{lstlisting}
This extension comes at an \textit{incosistency cost} of $7$. However, we now have the additional information that \texttt{action(give\_meds)} weighs $7$ in all extensions, while \texttt{action(dont\_give\_meds)} weighs $9$. Therefore, it is \textit{slightly} recommended \textit{not to give meds} in this scenario, thus respecting autonomy. This suggestion could of course be overridden by a human operator if s/he wishes to satisfy beneficence over autonomy.

We illustrated how wABA may resolve conflicts between ethical principles by deriving attacks from contraries and evaluating which sets of assumptions can be tolerated together under the given inconsistency budget. The weighted implementation allows discarding some attacks--e.g., between autonomy and beneficence--if the total weight of discarded attacks remains within the user-specified limit.

Note that this is particularly useful in this scenario as it also allows for reasoning about ethical principles in highly inconsistent scenarios, where plain ABA would not produce extensions, as it is often the case in realistic scenarios due to ethical principles leading to different courses of actions. In such cases wABA can help a human operator getting the fuller picture, e.g., verify what principles are satisfied in most extensions, normalizing their weight according to inconsistency levels, as well as showing what course of action is recommended and what this implies on the ethical dimension.
\end{example}

\section{Related Work}\label{related}

Resolving conflicts between ethical principles has long been a core challenge in both normative ethics and applied biomedical ethics.   
Clinical decision-making frequently necessitates the careful balancing of competing moral considerations, such as honoring patient autonomy, promoting well-being, preventing harm, and ensuring fairness. The principlist approach proposed by Beauchamp and Childress \cite{beauchamp2019principles} remains a cornerstone of biomedical ethics.  It articulates four central principles -autonomy, beneficence, nonmaleficence, and justice - that guide ethical decision-making. {In his influential work \cite{Floridi2023}, Floridi further argues that a fifth principle, namely \emph{explicability} is a required addition to those principles, specifically for the needs of AI ethics.}
However, these ethical frameworks do not prescribe a unique resolution when these principles come into conflict, prompting the need for formal methods that can assist in principled judgment.
Subsequent proposals, such as the mixed consequentialist-nonconsequentialist hierarchy, advocate balancing principles within categories (e.g., nonmaleficence vs. beneficence) before applying lexical priority to nonconsequentialist outcomes \cite{veatch1995resolving}. 
{Similarly, \cite{Townsend2022} provides a general workflow to derive concrete rules from general principles, taking into account the specificity of the contexts, and the allowable exceptions to rules. This approach has been subsequently implemented in Datalog \cite{Mirani2024}}.
These frameworks emphasize the role of case-based reasoning and iterative specification to resolve dilemmas.

Several computational approaches have been proposed to tackle this issue.
Anderson and Anderson, in their \textit{MedEthEx} \cite{AndersonAA05} and \textit{EthEl} \cite{AndersonA08}, operationalize Beauchamp and Childress' principles through machine learning, deriving decision rules from biomedical ethicists' intuitions in training cases. In these systems, the intensity of duty violations (e.g., autonomy vs. beneficence) is quantified by assigning it a weight, then, for each possible action, the system computes the weighted sum of duty satisfaction. After that, using inductive logic programming (ILP), their systems generate actionable rules for recurring dilemmas. 
In \cite{Rossi15} and \cite{LoreggiaMRV18} , Rossi and her team formalize principles as constraint-based systems with conflict resolution engines. They argue that preferences can model the relative importance of ethical principles and help resolve conflicts by identifying most preferred outcomes, given context-sensitive constraints. Their work leverages CP-nets and weighted constraints to evaluate ethical options, offering a flexible and explainable approach to value-sensitive decision-making.
In \cite{KLEIMANWEINER2017107}, Kleiman-Weiner et al. suggest an abstract and recursive utility calculus to resolve conflicts among moral principles. Moral theories (for the purposes of trading off different agents’ interests) can be
formalized as values or weights that an agent attaches to a set of abstract principles for how to factor any other agents’ utility functions into their own utility-based decision-making and judgment.
Drawing on machine learning and computational social choice, \cite{NoothigattuGADR18} proposes an algorithm to learn a model of societal preferences, and, when faced with specific ethical dilemma at runtime, aggregate those preferences to identify a desirable choice. 
Awad et al. in \cite{AWAD2020103349}, propose a framework for incorporating public opinion, as essential tool, into policy making in situations where ethical values are in conflict. Their framework advocates creating vignettes representing abstract value choices, eliciting the public’s opinion on these choices, and using machine learning, in particular ILP, to extract principles that can serve as succinct statements of the policies implied by these choices and rules that can be embedded in algorithms to guide the behavior of AI-based systems. To present
the functionality of this proposal, the authors borrow vignettes from the Moral Machine website \cite{awad2018moral}.

Dennis et al. \cite{DennisFSW16} developed the ETHAN (a BDI (Belief-Desire-Intention) agent language) system that deals with situations
when civil air navigation regulations are in conflict. The system relates these rules to four hierarchical ordered ethical principles (do not harm people, do not harm animals, do not damage self,
and do not damage property) and develops a course of action that generates the smallest violation to those principles in case of conflict. In their prototype, ethical reasoning was integrated into a BDI agent programming language via the agent plan selection mechanism.
\cite{NetoSL11} implement a BDI architecture within a multi-agent system (MAS), with a particular emphasis on handling norm conflicts. In their framework, agents are capable of adopting and dynamically updating norms, and they determine which norms to activate based on the current context, their desires, and their intentions. Conflicts between norms are resolved by selecting the norm that best contributes to the fulfillment of the agent’s goals and intentions, effectively embedding norm adherence within the agent’s motivational structure. Similarly, Mermet and Simon \cite{MermetS16} address norm conflicts by distinguishing between moral and ethical rules, the latter being invoked when moral rules are in conflict. They perform a verification of whether their system called GDT4MAS, is able to choose the correct ethical rule in conflict cases.

Chorley et al. in \cite{ChorleyBM06} described an implementation of the approach to deliberation about a choice of action based on presumptive argumentation and associated critical questions \cite{walton1996argumentation}. The authors use the argument scheme proposed in \cite{Atkinson05} to generate presumptive arguments for and against actions, and then subject these arguments to critical questioning.
They have explored automation of argumentation for practical reasoning by a single
agent in a multi-agent context, where agents may have conflicting values. Their approach was illustrated with a particular example based on an ethical dilemma.

{\cite{Bistarelli2018} uses abstract argumentation for decision-making with multiple experts. The approach represents in the form of  attacks in abstract argumentation, the lack of fairness in the evaluation performed by an expert, which may thus lead to the dismissal of the evaluation. }
{Finally, \cite{Liao2023} is directly related with our approach. It introduces a moral advisor based on logic-based normative systems integrated with Aspic-style structured  argumentation, in order to handle moral dilemmas involving multiple stakeholders, bearing different interests and ethical views.}



\section{Conclusion and future work}
We have introduced \textit{wABA}, together with an ASP based implementation. We applied our formalism to a patient dilemma scenario, showing how it can smoothly integrate the weighted assessment of a medical situation, the ethical principles involved, and possibly weighted solutions to the dilemma.  
We believe that this framework may provide a useful module for implementing reasoning with ethical principles in AI systems that interact with humans. 
Ongoing work is devoted to a detailed formal analysis of the framework, an extension of the implementation, and its applications in the ethical domain. Concerning the first aspect, we plan to analyze the computational properties of wABA and its implementation, such as correctness w.r.t. the intended semantics and analysis of its complexity. We would also like to further explore the relation to the different choices of the semantics and the choice of  semiring. We plan to compare our framework to other approaches, based on structured argumentation frameworks, in particular involving preferences, such as ASPIC+ \cite{VanEemeren2018} and ABA+ \cite{Toni2016}.

Concerning the implementation, we are incorporating further formal argumentation semantics and operations for the aggregation of weights. 

Finally, for applications, we are currently investigating realistic scenarios, involving reasoning with data and conflicting ethical principles, where we believe that our system may provide valuable support. A particularly promising application in this sense would be the analysis of medical triage, with their related intricate legal and ethical regulations \cite{Vantu2023}.

\begin{acknowledgments}
This work was partially supported by the project FAIR- Future AI Research (PE00000013), under the NRRP MUR program funded by the NextGenerationEU. 
\end{acknowledgments}    

\section*{Declaration on Generative AI}
  The authors have not employed any Generative AI tools.
  

\bibliography{bibliography}
\end{document}